\documentclass[runningheads]{llncs}
\usepackage[T1]{fontenc}
\usepackage{graphicx}
\usepackage{soul}

\usepackage{array}
\usepackage{amssymb}
\usepackage{amsmath} 
\usepackage{booktabs}
\usepackage{marvosym}
\usepackage{multirow}
\usepackage{amsfonts}
\usepackage{algorithm} 
\usepackage{geometry}
\usepackage{algpseudocode} 
\begin{document}
%

\title{PepEDiff: Zero-Shot Peptide Binder Design via Protein Embedding Diffusion}
%
%
\author{
Po-Yu, Liang\inst{1} \and Tibo, Duran\inst{2} \and Jun,  Bai\inst{1}\textsuperscript{\Letter}}
%
%
\institute{
    Department of Computer Science, University of Cincinnati, Ohio, United States\\
    \email{baiju@ucmail.uc.edu}
    \and
    Department of Microbiology and Plant Pathology, 
University of California, Riverside, California, United States
}
\maketitle              
\begin{abstract}

We present \textbf{PepEDiff}, a novel peptide binder generator that designs binding sequences given a target receptor protein sequence and its pocket residues. Peptide binder generation is critical in therapeutic and biochemical applications, yet many existing methods rely heavily on intermediate structure prediction, adding complexity and limiting sequence diversity. Our approach departs from this paradigm by generating binder sequences directly in a continuous latent space derived from a pretrained protein embedding model, without relying on predicted structures, thereby improving structural and sequence diversity. To encourage the model to capture binding-relevant features rather than memorizing known sequences, we perform latent-space exploration and diffusion-based sampling, enabling the generation of peptides beyond the limited distribution of known binders. This zero-shot generative strategy leverages the global protein embedding manifold as a semantic prior, allowing the model to propose novel peptide sequences in previously unseen regions of the protein space. We evaluate PepEDiff on TIGIT, a challenging target with a large, flat protein–protein interaction interface that lacks a druggable pocket. Despite its simplicity, our method outperforms state-of-the-art approaches across benchmark tests and in the TIGIT case study, demonstrating its potential as a general, structure-free framework for zero-shot peptide binder design.
The code for this research is available at GitHub\footnote{https://github.com/LabJunBMI/PepEDiff-An-Peptide-binder-Embedding-Diffusion-Model}.


\end{abstract}

\newpage
\section{Introduction}
Peptides serve critical functions across a wide range of applications, including fundamental biological research and the development of therapeutic strategies such as cancer immunotherapy\cite{wang2022therapeutic}. Conventional approaches to peptide discovery are typically labor-intensive, time-consuming, and costly\cite{hashemi2024therapeutic}. In recent years, deep learning methods have significantly transformed the design of peptides by allowing the generation of functional sequences in silico with reduced reliance on experimental screening. \cite{duran2024might} Early generative models, such as variational autoencoders and GANs, were applied to generate candidate sequences\cite{greener2018design,gupta2019feedback}. Recent large protein language-based methods enable sequence generation within learned embedding spaces, improving the functional relevance of designed peptides. \cite{zeinalipour2024design,luo2025cpl,jin2025ampgen}. Although computer-aided peptide discovery has been extensively studied\cite{sharma2023peptide}, the direct receptor-specific design of peptide inhibitors remains a less-developed area.\cite{wan2022deep} Structure-guided approaches, such as RFdiffusion \cite{watson2022broadly}, synthesize peptide backbones around receptor pockets and subsequently recover sequences using inverse folding models like ProteinMPNN \cite{dauparas2022robust}. DiffPepBuilder \cite{wang2024target}, in contrast, adopts a joint optimization strategy, simultaneously designing both structure and sequence to better accommodate the binding interface. Despite their success, structure-based peptide design methods face notable limitations. By relying heavily on intermediate structural representations, these approaches often generate peptides dominated by $\alpha$-helical conformations (as shown in our evaluation results, Figure~\ref{fig:tigit_performance_peptide}), which restricts structural diversity. Furthermore, misalignment between generated structures and their inferred sequences can introduce cascading errors \cite{ye2024proteinbench,xue2025improving}. The relatively small number of available peptide binders compared to the vast protein space also constrains the diversity of generated binders in both structural and sequence space.

These limitations are particularly evident in challenging targets such as TIGIT, an immune checkpoint receptor characterized by a large, flat protein–protein interaction (PPI) interface and high binding affinity for its ligand, PVR (CD155)\cite{cui2025tigit}. Such interfaces lack well-defined druggable pockets, rendering traditional small-molecule inhibitors ineffective\cite{arkin2014small}. Although several anti-TIGIT monoclonal antibody (mAb) programs have been developed, many have failed to meet primary endpoints in overall survival or exhibited increased adverse effects\cite{NCT04738487,NCT05226598,NCT04294810,NCT04619797}. 
Peptides, occupying a middle ground between antibodies and small molecules, offer promising potential for modulating such “undruggable” PPIs\cite{wang2021rational}. 

To address these challenges, we propose a novel framework that eliminates the need for intermediate structure prediction. Our approach designs peptides directly within a learned embedding space, enabling the generation of sequences that preserve key functional properties without being limited by predefined sequence or structural templates. Additionally, we incorporate a latent-space exploration strategy inspired by prior work \cite{liang2024exploring} to expand sampling beyond the distribution of known peptide binders. This combination not only yields binders with improved affinity but also enhances the diversity of generated peptides, unlocking candidates beyond the scope of existing methods. In summary, our contributions include: \textbf{1)} Proposed a structure-independent peptide binder embedding design framework that operates solely from sequence and pocket residue information. \textbf{2)} Proposed a zero-shot generation framework, demonstrating enhanced diversity alongside superior binding performance compared to state-of-the-art methods on the testing set. \textbf{3)} Validated our approach through a case study using physics modeling on designing binders for the TIGIT (UniProt ID: $Q495A1$ \cite{uniprot2025uniprot}) receptor.
\section{Method}
\subsection{Problem Formulation}
\begin{figure}[htbp]
    \vspace{-0.1 in}
    \centering
    \includegraphics[width=\columnwidth]{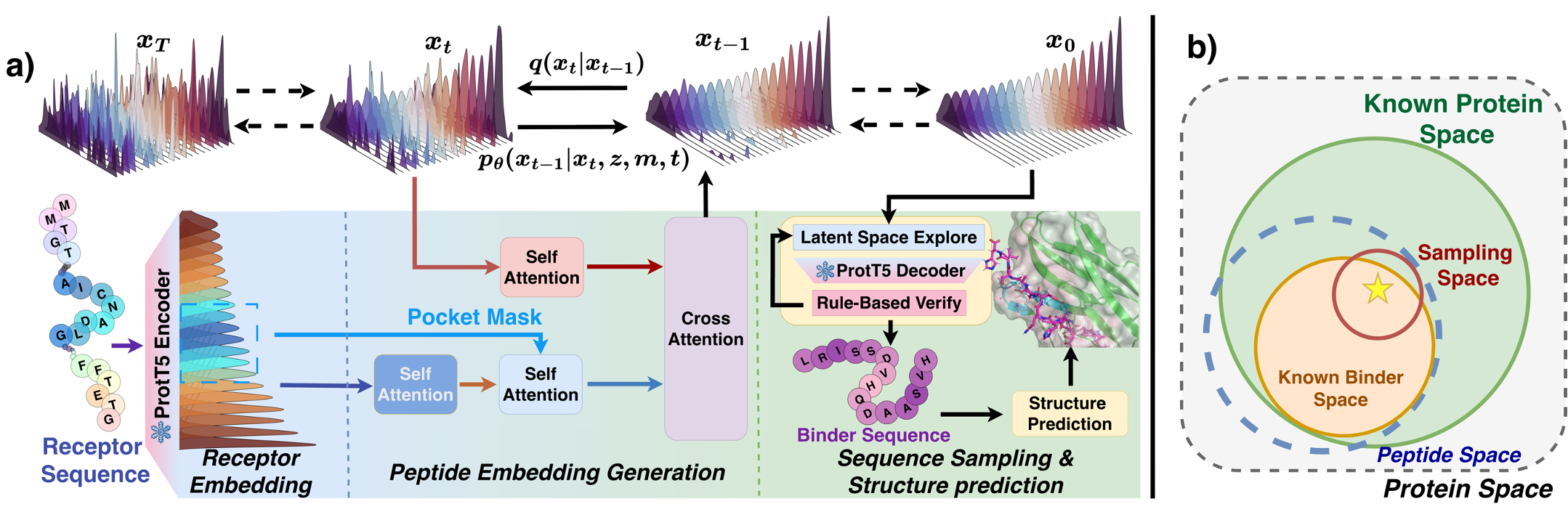}
    \caption{
    \textbf{Overview of the structure-free peptide design framework.} \textbf{a) }We adopt the diffusion framework (top row), which iteratively refines a random signal ($x_T$) into a final peptide embedding ($x_0$). The reverse denoising process, $p_\theta(x_{t-1}|x_t, z, m)$, is conditioned on the target receptor at each step. The denoising network (bottom row) employs a cross-attention mechanism to focus on the receptor's binding pocket, represented by its embedding $z$ and a pocket mask $m$. The final refined embedding $x_0$ is then decoded to generate the binder sequence $\hat{S}$. The generated sequence is subsequently used to predict the receptor–peptide complex structure. \textbf{b)} An illustration of out-of-distribution sampling. By exploring the latent space of all know protein sequence, we aim to generate peptides out of the relative small know peptide binder distribution.}
    \label{fig:method_flow}
    \vspace{-0.1 in}
\end{figure}
Let $\mathcal{A}$ denote the set of 20 standard amino acids. Given a target receptor protein sequence $r \in \mathcal{A}^L$ and a binary mask $m \in \{0, 1\}^L$ indicating pocket residues, where the $L$ is the length of receptor sequence, the goal is to generate a peptide sequence $\hat{s} \in \mathcal{A}^{L'}$, where $L'$ is the peptide sequence length, that effectively binds to the specified receptor pocket. To achieve this, we first encode the receptor sequence into a continuous representation using an encoder network $z = \text{Encoder}(s)$, where $z \in \mathbb{R}^d$ serves as the conditioning context. We then employ a conditional diffusion model\cite{ho2020denoising} to generate the peptide in a learned embedding space. Specifically, we initialize the peptide embedding as Gaussian noise $x_T \sim \mathcal{N}(0, I)$, and iteratively denoise it using a neural network conditioned on receptor embedding, $z$, and the denoising timestep $T$:
\[
x_{t-1} \sim p_\theta(x_{t-1} \mid x_t, z, m, t), \quad \text{for } t = T, T-1, \dots, 1
\]
The final denoised embedding $x_0$ represents the peptide candidate in latent space, which is subsequently decoded into an amino acid sequence using a decoder network $\hat{s} = \text{Decoder}(x_0)$. In our implementation, we employ the ProtT5\cite{elnaggar2021prottrans} embedding model, whose encoder–decoder architecture is well-suited to our proposed framework. Both the encoder and decoder are kept frozen during both training and inference. Beyond this primary generative pathway, we introduce a complementary zero-shot latent-space exploration mechanism designed to access binding-relevant regions that lie outside the empirical peptide distribution. This is achieved by perturbing embeddings of known peptides using scaled Gaussian noise, enabling the model to sample around the observed embeddings and explore points in the broader binding-relevant manifold. By operating directly in the continuous embedding space and conditioning on receptor-specific information, the method can generate peptide binders without requiring structural intermediates and without relying on examples from the unseen regions it ultimately discovers, thereby enabling true zero-shot peptide binder generation.



\subsection{Diffusion Process}

We adopt a denoising diffusion probabilistic model (DDPM)\cite{ho2020denoising} to generate peptide embeddings in a stepwise manner. As illustrated in Figure~\ref{fig:method_flow}A, diffusion models consist of a forward noising process, $q$, which gradually corrupts the data by adding Gaussian noise, and a reverse denoising process, $ p_\theta$, which learns to recover the original data from these noisy representations.

\paragraph{Forward Noising Process} 
Let  $x_0 \in \mathbb{R}^{L' \times d}$ denote the clean peptide embedding, where $L'$ is the peptide length and $d$ is the embedding dimension. The forward process defines a Markov chain that corrupts the data at each step $t$ by adding Gaussian noise:

\begin{equation}
q(x_t \mid x_{t-1}) = \mathcal{N}(x_t; \sqrt{\alpha_t} x_{t-1}, (1 - \alpha_t) \mathbf{I})
\end{equation}

where the noise schedule \( \{\alpha_t\} \) is derived from a cosine-based function\cite{nichol2021improved}:

\begin{equation}
\bar{\alpha}_t = \left( \frac{\cos\left( \frac{t/T + s}{1 + s} \cdot \frac{\pi}{2} \right)}{\cos\left( \frac{s}{1 + s} \cdot \frac{\pi}{2} \right)} \right)^2
\end{equation}

with with a small offset $s$ for numerical stability and $ \alpha_t = \bar{\alpha}_t / \bar{\alpha}_{t-1} $. This yields the marginal distribution:

\begin{equation}
q(x_t \mid x_0) = \mathcal{N}(x_t; \sqrt{\bar{\alpha}_t} x_0, (1 - \bar{\alpha}_t) \mathbf{I}).
\end{equation}

\paragraph{Reverse Denoising Process}

The reverse process learns to iteratively remove noise from $x_t$, conditioned on the receptor embedding $z$, pocket mask $m$, and timestep $t$. It is modeled as a Gaussian distribution with a learned mean:

\begin{equation}
p_\theta(x_{t-1} \mid x_t, z, m, t) = \mathcal{N}\big(x_{t-1}; \mu_\theta(x_t, z, m, t), \sigma_t^2 \mathbf{I}\big),
\end{equation}

where $\sigma_t^2$ is a fixed variance from the noise schedule. During training, the neural network is parameterized as $\epsilon_\theta(x_t, z, m, t)$ and is trained to predict the noise vector $\epsilon$ used to generate $x_t$. At inference time, this prediction is used to compute the mean $\mu_\theta$ according to the DDPM formulation:

\begin{equation}
\mu_\theta(x_t, z, m, t) = \frac{1}{\sqrt{\alpha_t}} \left(x_t - \frac{1 - \alpha_t}{\sqrt{1 - \bar{\alpha}_t}} \epsilon_\theta(x_t, z, m, t) \right),
\end{equation}

where $\alpha_t$ and $\bar{\alpha}_t$ are the noise schedule coefficients. A sample is drawn by adding noise to this mean:

\begin{equation}
x_{t-1} = \mu_\theta(x_t, z, m, t) + \sigma_t \epsilon, \quad \epsilon \sim \mathcal{N}(0, \mathbf{I}).
\end{equation}

This formulation allows the model to reverse the corruption process and generate clean peptide embeddings conditioned on receptor and pocket information.

\subsection{Model Architecture \& Loss Terms}
As shown in Figure~\ref{fig:method_flow}B, our model consists of several attention-based\cite{vaswani2017attention} modules designed to parameterize the denoising function $\mu_\theta(x_t, z, m, t)$. The receptor sequence is first encoded using a pretrained ProtT5 model, yielding contextual embeddings $z \in \mathbb{R}^{L \times d}$. A self-attention layer refines $z$, followed by another self-attention layer modulated by a binary pocket mask $m \in \{0,1\}^L$, producing the pocket-specific representation $z_{\text{pocket}} \in \mathbb{R}^{L \times d}$. The noised peptide embedding $x_t \in \mathbb{R}^{L' \times d}$ is processed through a self-attention layer to capture internal dependencies, and then updated via a cross-attention layer using the pocket representation as context. More detailed configuration can be found in both the GitHub repository and section 1 of Appendix.

To train the model, we follow the standard DDPM objective by predicting the noise vector added during the forward process. Given a clean embedding $x_0$ and its noisy version $x_t$, generated as
\[
x_t = \sqrt{\bar{\alpha}_t} x_0 + \sqrt{1 - \bar{\alpha}_t} \epsilon, \quad \epsilon \sim \mathcal{N}(0, \mathbf{I}),
\]
the model $\epsilon_\theta(x_t, z, m, t)$ is trained to approximate $\epsilon$.

The first loss term is the mean squared error (MSE) between the predicted and true noise:
\begin{equation}
\mathcal{L}_{\text{MSE}} = \| \epsilon_\theta(x_t, z, m, t) - \epsilon \|_2^2.
\end{equation}

The second term is a cosine similarity loss computed over residue-wise noise vectors:
\begin{equation}
\mathcal{L}_{\text{cos}} = 1 - \frac{1}{L'} \sum_{i=1}^{L'} \frac{\epsilon_\theta^{(i)} \cdot \epsilon^{(i)}}{\|\epsilon_\theta^{(i)}\|_2 \cdot \|\epsilon^{(i)}\|_2},
\end{equation}
where $\epsilon^{(i)}$ and $\epsilon_\theta^{(i)}$ are the $i$-th row vectors of the true and predicted noise, respectively.

The final training loss combines both terms:
\begin{equation}
\mathcal{L} = \lambda_{\text{MSE}} \mathcal{L}_{\text{MSE}} + \lambda_{\text{cos}} \mathcal{L}_{\text{cos}},
\end{equation}
with $\lambda_{\text{MSE}} = 0.9$ and $\lambda_{\text{cos}} = 0.1$ in our implementation.

\subsection{Zero-shot generative exploration of peptide space.}
Due to the limited number of available peptide binders ($N = 4{,}758$; see Section~\ref{sec:dataset}), we designed a zero-shot latent-space exploration framework that leverages a pretrained protein embedding model to generate peptide candidates beyond the training distribution.

As illustrated in Figure~\ref{fig:method_flow}b, the latent space (green region) contains the global protein manifold, within which the \textbf{known peptide binders} (orange region) form only a small subspace. We denote by
\[
\mathcal{X}_{\text{bind}} \subseteq \mathcal{X}_{\text{protein}}
\]
the broader \textbf{binding-relevant region} (dashed blue and red overlap), which consists of embedding locations that are likely relevant for peptide--receptor binding but are not necessarily observed in the data. The embeddings of the observed peptide binders are
\[
\mathcal{X}_{\text{peptide}} = f(S_{\text{peptide}}) \subsetneq \mathcal{X}_{\text{bind}}.
\]
The \textbf{unseen region} that our method aims to explore is therefore defined as
\[
\mathcal{X}_{\text{unseen}}
=
\mathcal{X}_{\text{bind}} \setminus \mathcal{X}_{\text{peptide}},
\]
representing binding-relevant embedding space that is not covered by the training set. Our objective is to sample within this unseen region while remaining grounded in the global protein embedding distribution.

To reach such out-of-distribution locations, we perturb each peptide embedding $x_i \in \mathcal{X}_{\text{peptide}}$ using Gaussian noise:
\[
x_i' = x_i + \sigma \epsilon, \qquad \epsilon \sim \mathcal{N}(0, I_d),
\]
where the scale parameter $\sigma$ controls the radius of exploration. Each perturbed point $x_i'$ is decoded by the pretrained ProtT5 decoder to obtain a candidate peptide sequence. For each $x_i$, fifty decoding attempts are performed at an initial scale $\sigma = 0.3$; if no valid sequence is obtained, $\sigma$ is increased by $0.1$ to enable progressively broader exploration of the protein manifold.

This latent-space perturbation enables \emph{zero-shot generation} by exploiting the pretrained protein manifold as a semantic prior. Although the generative model is trained solely on the limited peptide dataset $S_{\text{peptide}}$, sampling within the globally informed space $\mathcal{X}_{\text{protein}}$ allows it to synthesize binding-relevant peptides in previously unseen regions, thereby performing \emph{out-of-distribution discovery} guided by protein-level sequence knowledge.

When transforming embeddings back into sequence space, we observed that the generated peptides often exhibited two types of artifacts: (i) dominance of a single amino acid type, or (ii) long homogeneous motifs composed of repeated residues. These artifacts likely arise from sampling regions of the latent space that are sparsely populated or insufficiently supported by the decoder. To mitigate this issue, we applied two filtering criteria. A generated sequence was discarded if: (1) more than $50\%$ of its residues were identical, or (2) any contiguous segment of identical residues exceeded $30\%$ of the total sequence length. These thresholds help ensure a more effective and biologically meaningful exploration of the peptide embedding space.

\section{Data \& Experiment Setup}

\subsection{Dataset \& Pre-Processing}\label{sec:dataset}

We use the open-source BioLip database \cite{zhang2024biolip2} to evaluate our method. BioLip contains a large collection of protein-ligand interactions involving various ligand types, including peptides in its most recent release. Out of $781,684$ total entries, $35,167$ are protein-peptide interactions. We remove entries with duplicated PDB IDs and exclude any structures with a resolution worse than 5~\AA{} to ensure compatibility with other structure-based models, resulting in $14,748$ high-quality records. To ensure the completeness of peptide binder sequences, we retrieve full sequences from the RCSB Protein Data Bank\cite{berman2000protein}, rather than using those provided in the structure files, which may be truncated. Entries containing unknown residues are discarded, leaving $11,984$ clean records.

Despite removing duplicated PDB IDs, potential data leakage could still occur due to homologous receptors. To mitigate this, we apply MMSeqs2\cite{steinegger2017mmseqs2} clustering using a minimum sequence identity of $50\%$ and at least $80\%$ alignment coverage. To prevent any single cluster from dominating the dataset, we retain a maximum of $10$ randomly selected sequences per cluster. For the test set, we randomly sample $5\%$ of the clusters and select one representative sequence from each cluster, resulting in a test set of $311$ entries. The remaining clusters are divided into training and validation sets using an $80:20$ split at the cluster level. Due to variability in cluster sizes, the final split contains $4,758$ training records and $546$ validation records.

\subsection{Baseline Model \& Evaluation Metrics}
We compare our method with two state-of-the-art baselines that represent the two dominant strategies for peptide binder design. The first baseline uses the well-known RFDiffusion\cite{watson2022broadly} model to design binder backbone structures. Combined with an inverse folding model, ProteinMPNN\cite{dauparas2022robust}, it yields both sequence and structure for each generated peptide binder. In contrast, the second baseline, DiffPepBuilder\cite{wang2024target}, follows a co-design strategy, simultaneously generating both sequence and structure of the binder in a unified model.

To assess performance, we evaluate all methods using four key metrics: \textbf{sequence diversity}, \textbf{structure diversity}, \textbf{embedding diversity} and \textbf{binding energy}. For our method and DiffPepBuilder, we generate $10$ binders per receptor. For the RFDiffusion and ProteinMPNN (RF\&MPNN) pipeline, we first generate $10$ backbone structures using RFDiffusion and then generate one sequence for each using ProteinMPNN, yielding a total of $10$ binders, comparable to the other two methods.

\paragraph{Sequence Diversity}
To assess the diversity of generated sequences, we calculate pairwise differences between amino acid sequences. Since some amino acid substitutions are more similar biologically than others, we use a substitution matrix to weight differences accordingly. Specifically, we adopt the widely used BLOSUM62\cite{henikoff1992amino} matrix and compute similarity scores using the Needleman-Wunsch (NW) global alignment algorithm\cite{needleman1970general}. The normalized pairwise sequence similarity is defined as:

\begin{equation}
\text{Sim}_{\text{seq}}(s_1, s_2) = \frac{\text{NW}(s_1, s_2)}{\text{NW}(s_1, s_1)}
\end{equation}

where $s_1$ and $s_2$ are two sequences being compared, and $\text{NW}(s_1, s_1)$ represents the self-alignment score used for normalization. This ensures the similarity score is independent of sequence length and the absolute substitution scores.

Using this similarity measure, we defined sequence diversity as the average of similarity complements over all unique pairs in a set:

\begin{equation}
\text{Div}_{\text{seq}}(S) = \frac{1}{N(N-1)} \sum_{i}^N \sum_{j \neq i}^N (1-\text{Sim}_{\text{seq}}(s_i, s_j))
\end{equation}

where $S = \{s_1, s_2, \dots, s_N\}$ is a set of $N$ sequences.

\paragraph{Structure diversity}
Similar to sequence diversity, we measure structural diversity by first computing pairwise structure similarity using the TM-score\cite{zhang2004scoring}. The TM-score is a length-normalized metric bounded between 0 and 1, with higher scores indicating greater similarity. Unlike root mean square deviation (RMSD), TM-score emphasizes global structural similarity and is less sensitive to local deviations. Structure diversity is then defined as:

\begin{equation}
\text{Div}_{\text{str}}(U) = \frac{1}{N(N-1)} \sum_i^N \sum_{j \neq i}^N ({1 - \text{TM}(u_i, u_j)})
\end{equation}

where $U = {u_1, u_2, \dots, u_N}$ is a set of $N$ predicted structures.

\paragraph{Embedding Similarity \& diversity} To assess whether the generated embeddings resemble the ground truth while remaining diverse among themselves, we define two metrics: embedding similarity and embedding diversity.

To compute the similarity between protein sequence embeddings, we first average the embedding matrix of shape $(L+1) \times 1024$ across the sequence length to obtain a single vector representation of shape $1024$. The similarity between two such embeddings is then calculated using cosine similarity:
\begin{equation}
\text{Sim}_{\text{cos}}(e_1, e_2) = \frac{e_1 \cdot e_2}{\lVert e_1 \rVert \lVert e_2 \rVert}
\end{equation}

where $e_1, e_2 \in \mathbb{R}^{1024}$ are the mean embeddings of two sequences.

Based on this, we define embedding diversity for a set of embeddings $E = \{e_1, e_2, \dots, e_N\}$ as the average pairwise dissimilarity:
\begin{equation}
    \text{Div}_{\text{emb}}(E) = \frac{1}{N(N-1)} \sum_i^N \sum_{j \neq i}^N ({1 - \text{Sim}_{cos}(u_i, u_j)})
\end{equation}

\paragraph{Rosetta Estimated Energy}\label{sec:rosetta_estimated_energy}
To assess binding affinity, we estimate the binding energy between the designed peptide binder and the receptor using the Rosetta Commons toolkit\cite{chaudhury2010pyrosetta,leaver2013scientific}. All structures are first relaxed using the FastRelax protocol\cite{simons1999improved} to ensure compatibility with the Rosetta energy function. The binding energy is then computed as:

\begin{equation}
\Delta G(u_c) = \text{Energy}(u_c) - \left[ \text{Energy}(u_r) + \text{Energy}(u_s) \right],
\end{equation}

where $u_c$ denotes the full complex, and $u_r$, $u_s$ refer to the receptor and binder structures, respectively.

For baseline methods that generate both the binder structure and its binding pose (e.g., RF\&MPNN and DiffPepBuilder), we directly use the predicted conformations. Since DiffPepBuilder outputs only the pocket region, we align the generated binder back to the full receptor structure to enable consistent evaluation. Additionally, both baseline methods produce only backbone coordinates, without explicit side-chain atoms. To complete these structures, we use the PackRotamers\cite{renfrew2014rotamer} protocol from the Rosetta Commons toolkit to reconstruct side chains. For our method, which predicts only the peptide sequence, we employ the Boltz-2\cite{passaro2025boltz} structure prediction model to recover the binder conformation and binding pose, conditioned on both the receptor and pocket information.

After relaxation, we verify whether the predicted binder remains in contact with the pocket. Complexes without any binder-pocket contact, defined as a minimum atom distance exceeding 5~\AA{}\cite{salamanca2017optimal}, are considered non-binders and assigned a binding energy of $\Delta G(u_c) = 0$.
\vspace{-0.15 in}

\subsection{Case Study \& Physics Modeling Validation Setup}\label{sec:method_tigit}

\paragraph{Binder Generation for TIGIT} 
To evaluate our method in a practical setting, we conducted a case study on the immune receptor, T-cell immunoreceptor with immunoglobulin and ITIM domains (TIGIT), which currently has no well-characterized peptide binders. The crystal structure of TIGIT are obtained from RCSB PDB\cite{stengl2010crystal}(PDB ID: 3Q0H\cite{PDB_3Q0H}). The functional residues of TIGIT (A67–G74) were provided as binding pocket information during generation. We used our proposed method and two baselines to generate $100$ candidate binders each, with a fixed sequence length of $15$ residues. For the RF\&MPNN baseline, we first generated $10$ unique structural backbones and then sampled $10$ sequences for each backbone. In contrast, both DiffPepBuilder and our method directly produced $100$ unique peptide sequences. To identify the most promising candidates, we estimated the binding energy of all generated peptides using the Rosetta Commons toolkit. 
The top binder from each method was then selected for detailed analysis via molecular dynamics (MD) simulations. Generated peptide sequences are available in GitHub repository.


For each model, we selected the binder with the highest predicted binding energy difference and conducted a MD simulation using GROMACS (version 2023.4)\cite{abraham2015gromacs} with Optimized Potentials for Liquid Simulations All-Atom (OPLS-AA) force field\cite{jorgensen1996development} to evaluate its stability and interaction with the receptor. All simulations were conducted at 300 K in an extended simple point charge (SPC/E)\cite{berendsen1987missing,chatterjee2008computational} water solvent with NaCl ions at neutral pH to mimic the physiological condition. Umbrella sampling\cite{torrie1977nonphysical} was employed to estimate the free binding energy. 

The simulations were carried out in three stages. First, to obtain the initial peptide structure, we used AlphaFold2 \cite{jumper2021highly}, followed by three independent 1000 ns MD simulations. From each simulation, the final frame was extracted, and pairwise root mean square deviation (RMSD) values were calculated among the three structures. The structure showing the smallest overall RMSD compared to the others was selected as the representative conformation for subsequent simulations. Second, we built a system containing experiment method obtained TIGIT structure\cite{stengel2012structure} and the selected peptide structure. The peptide was positioned near the binding pocket residues, with its center of mass placed approximately 30 Å from the pocket center. Three independent dynamic docking MD simulations were then performed. Finally, the replicate showing stable interactions between the peptide and the binding pocket residues was analyzed using umbrella sampling, with weighted histogram analysis method (WHAM)\cite{kumar1992weighted} algorithm to compute the interaction free energies and analyze the trajectories. More Detailed simulation procedures and setup are provided in the Section 2 of Appendix.

\section{Result}
\vspace{-0.1 in}
\subsection{Overall Result}
\begin{table}[htbp]
\vspace{-0.3 in}
\centering
\caption{
Performance comparison between our model and two baseline methods.  \textbf{Testing set} reports results on the full testing set, while \textbf{TIGIT} focuses on binders generated for the immunoreceptor TIGIT. All values are reported as "mean (std)". Best scores are highlighted in \textbf{bold}.
}
{\begin{tabular}{l|l|r|r|r}

\multicolumn{2}{c|}{}& RF\&MPNN & DiffPepBuilder & Our Model \\
\hline

\multirow{3}{*}{\begin{tabular}{@{}l@{}} Testing Set\end{tabular}}

    &($\uparrow$)$\text{Div}_{\text{seq}}$ 
    &0.56\scriptsize{(0.09)} 
    &0.44\scriptsize{(0.09)} 
    &\textbf{0.67\scriptsize{(0.03)}} \\
    
    &($\uparrow$)$\text{Div}_{\text{str}}$ 
    &0.45\scriptsize{(0.21)} 
    &0.54\scriptsize{(0.17)} 
    &\textbf{0.72\scriptsize{(0.15)}} \\
    
    &($\downarrow$)$\Delta G$
    &-67.99\scriptsize{(35.04)} 
    &-45.51\scriptsize{(18.69)} 
    &\textbf{-78.34\scriptsize{(72.82)}} \\
\hline

\multirow{3}{*}{\begin{tabular}{@{}l@{}} TIGIT\end{tabular}}

    &($\uparrow$)$\text{Div}_{\text{seq}}$ 
    &0.45\scriptsize{(0.13)} 
    &0.39\scriptsize{(0.13)} 
    &\textbf{0.69\scriptsize{(0.09)} }
    \\

    &($\uparrow$)$\text{Div}_{\text{str}}$ 
    &0.14\scriptsize{(0.05)} 
    &0.46\scriptsize{(0.11)} 
    &\textbf{0.80\scriptsize{(0.10)}} \\
    
    &($\downarrow$)$\Delta G$
    &-28.62\scriptsize{(4.34)} 
    &-20.02\scriptsize{(4.86)} 
    &\textbf{-30.49\scriptsize{(6.97)}}
\end{tabular}}\label{tab:performance}
\end{table}
The performance comparison between our proposed method and the two baselines is summarized in Table~\ref{tab:performance}. In terms of sequence diversity, our model achieves the highest score ($0.67$), followed by RF\&MPNN ($0.56$), while DiffPepBuilder performs the worst ($0.44$). For structure diversity, DiffPepBuilder attains a moderate score of $0.54$, slightly higher than RF\&MPNN ($0.45$), whereas our model again outperforms both with a score of $0.72$. Regarding the estimated binding energy ($\Delta G$), our method demonstrates the strongest binding affinity, with an average $\Delta G$ of $-78.34$, compared to $-67.99$ for RF\&MPNN and $-45.51$ for DiffPepBuilder.

\begin{figure}[htbp]
    \vspace{-0.1 in}
    \centering
    \includegraphics[width=\columnwidth]{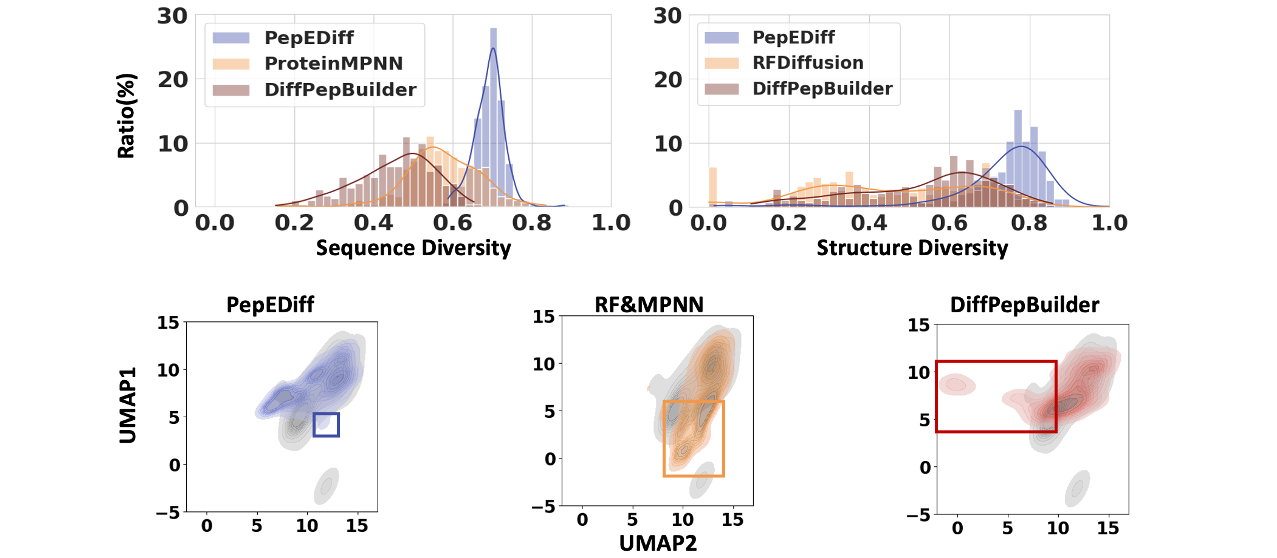}
    \caption{\textbf{Performance comparison with baseline methods.} Top row shows the sequence and structure diversities. Our model achieves superior diversity in both sequence and structure comparing to the two baseline model. Bottom row shows the distribution of generated peptides embedding (dimension reduced using UMAP\cite{mcinnes2018umap}) comparing to the testing set (shown as gray area). Major distribution differences are marked with color box.}
    \label{fig:general_performance}
    \vspace{-0.2 in}
\end{figure}
To evaluate whether the model captures functional properties beyond mere sequence similarity, we analyzed both embedding similarity and embedding diversity among the generated binders. As shown in Table~\ref{tab:performance}, the embedding similarity across the three methods is comparable, with RF\&MPNN slightly outperforming the others by a margin of $0.01$ (standard deviation: $0.22$). Notably, our model exhibits significantly higher embedding diversity ($0.41$) than both DiffPepBuilder ($0.21$) and RF\&MPNN ($0.27$), with p-values of $1.09 \times 10^{-68}$ and $7.61 \times 10^{-45}$, respectively. These results suggest that our method not only captures meaningful binder representations but also explores a broader functional space, generating diverse binders with similar binding properties rather than relying solely on sequence similarity.

\begin{figure}[htbp]
    \vspace{-0.2 in}
    \centering
    \includegraphics[width=\columnwidth]{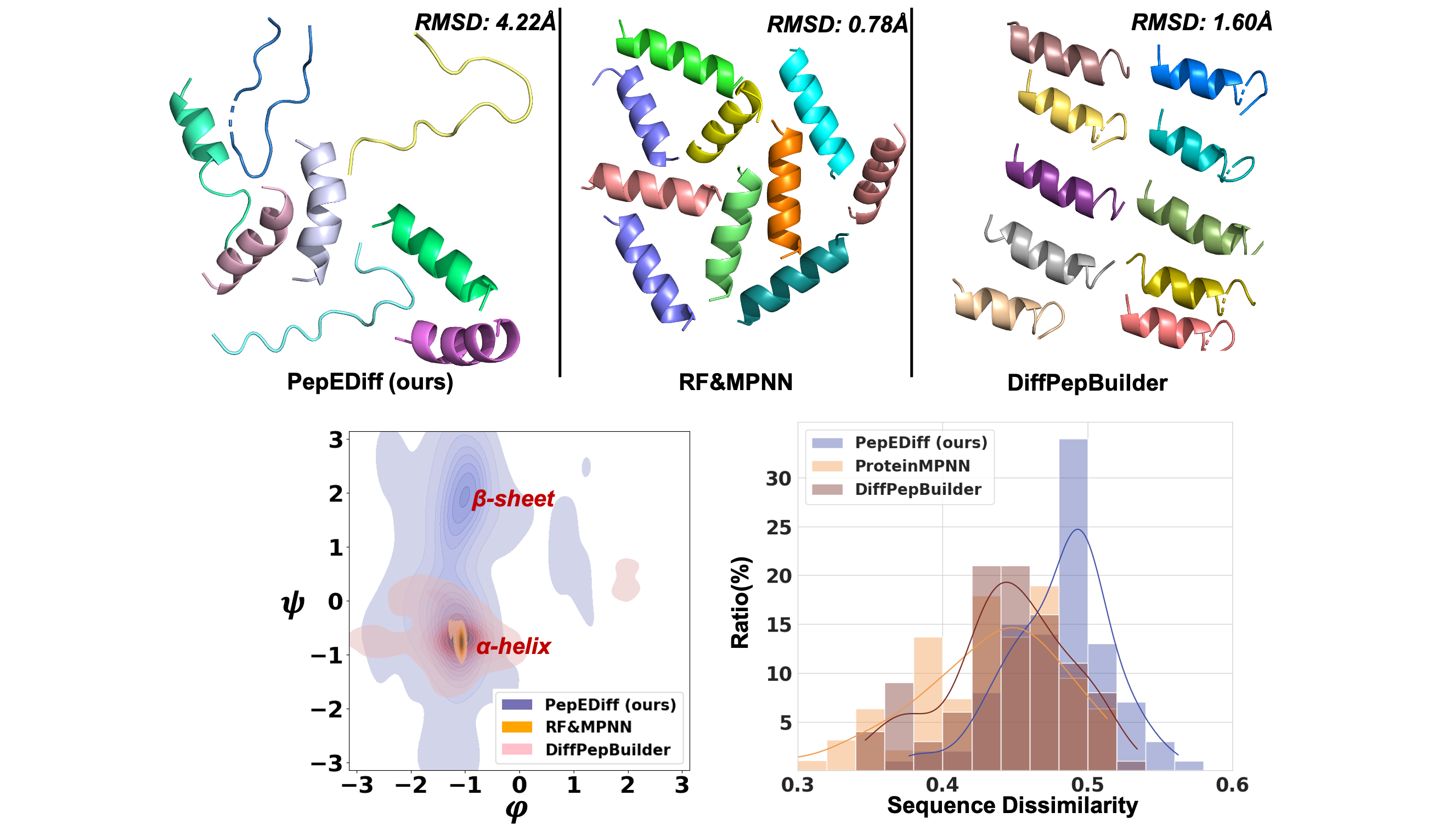}
    \caption{\textbf{Generated TIGIT binders.} 
    The top row shows some generated peptides for all baseline method. For the RF\&MPNN pipeline, all ten generated structures are visualized. For PepEDiff and DiffPepBuilder, ten peptides (out of 100) are shown for visual clarity. The bottom row shows a combined Ramachandran plot for all generated peptide structures, and the dissimilarity ($1-\text{similarity}$) between generated peptide and training set.}
    \label{fig:tigit_performance_peptide}
    \vspace{-0.2in}
\end{figure}
Notably, the standard deviation of $\Delta G$ for our model is considerably larger than that of RF\&MPNN and DiffPepBuilder, suggesting a broader range of binding strengths. To better understand these differences, we visualize the full distributions of sequence and structure diversity in the top row of Figure~\ref{fig:general_performance}. Both RF\&MPNN and DiffPepBuilder exhibit relatively lower sequence diversity compared to our model, whereas in terms of structural diversity, DiffPepBuilder shows a distribution more similar to ours than RF\&MPNN. This pattern, when considered alongside $\Delta G$, indicates that the broader sequence space explored by our model enables the discovery of stronger binders beyond the limited distribution of known peptide binders. The bottom row of Figure~\ref{fig:general_performance} shows that our model exhibits the smallest difference to the testing set embedding (background gray area). This observation further demonstrates that designing peptides in a continuous latent space allows our model to better capture true peptide properties while maintaining high sequence and structural diversity.
\vspace{-0.1 in}

\subsection{Binder Design for TIGIT}
To evaluate our method on a practical target, we conducted a case study generating 100 peptide binders for the immune receptor TIGIT. The diversity analysis, presented in table~\ref{tab:performance}, shows that our model achieved higher diversity compared to the baselines. For sequence diversity, RF\&MPNN and DiffPepBuilder yielded average scores of $0.45$ and $0.39$, respectively (both with a standard deviation of $0.13$), whereas our method reached an average of $0.69$. A similar trend was observed in structural diversity, where our model obtained the highest average score of $0.80$. In contrast, RF\&MPNN showed limited structural diversity, with an average of only $0.14$, likely due to its reliance on just $10$ unique backbones, compared to the $100$ distinct peptides generated by both DiffPepBuilder and our method. Figure~\ref{fig:tigit_performance_peptide} shows the Ramachandran plot of generated peptide structures. While RF\&MPNN and DiffPepBuilder primarily generated $\alpha$-helix structures, our method shows remarkable ability to explore wider range of peptide structures, including $\beta$-sheet region. To evaluate the zero-shot generation ability of our model, we plot the the sequence dissimilarity between generated peptides and the peptide in training set. Our model generated peptide sequences significantly more dissimilar to the training set comparing to RF\&MPNN and DiffPepBuilder (with p-value of $4.93*10^{-13}$ and $6.54*10^{-10}$).

\begin{figure}[htbp]
    \vspace{-0.3in}
    \centering
    \includegraphics[width=\columnwidth]{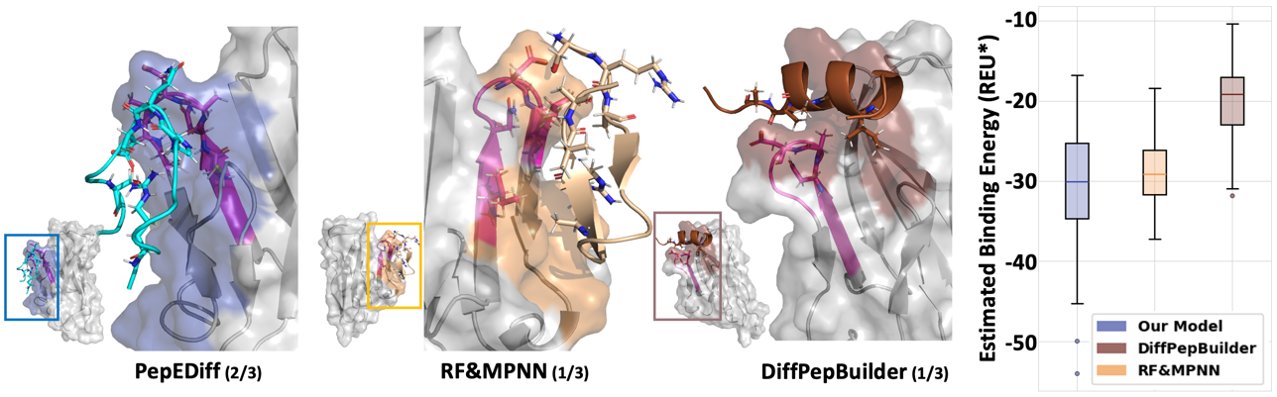}
    \textit{*REU}: Rosetta Energy Unit, a virtual energy unit used in the Rosetta Commons toolkit.
    \caption{\textbf{Top-Performing Binders for the TIGIT Receptor.} The three leftmost figures show the docking poses. The number of repeats showing interactions between the peptide and pocket is marked beside the method. The rightmost figure shows the energy distribution of generated peptides. The exact sequence of top-performing binder for our model, RF\&MPNN, and DiffPepBuilder are $LRISSDVHQDAASVH$, $SRAEQNAALLARVAG$, and $ILDDILRAAALAAGF$, respectively. All 100 generated binders are provided in our GitHub repository.
    }
    \label{fig:tigit_performance_str}
    \vspace{-0.23in}
\end{figure}

Binding affinity analysis, shown in the right panel of Figure~\ref{fig:tigit_performance_str}, further highlights the advantages of our approach. DiffPepBuilder showed the lowest performance. While our method and RF\&MPNN had comparable median binding energies, our model identified a broader set of peptides with strong predicted affinities. As a result, it achieved the lowest overall average binding energy of $-30.49$, compared to $-28.62$ for RF\&MPNN and $-20.02$ for DiffPepBuilder.

\begin{figure}[htbp]
    \centering
    \includegraphics[width=\columnwidth]{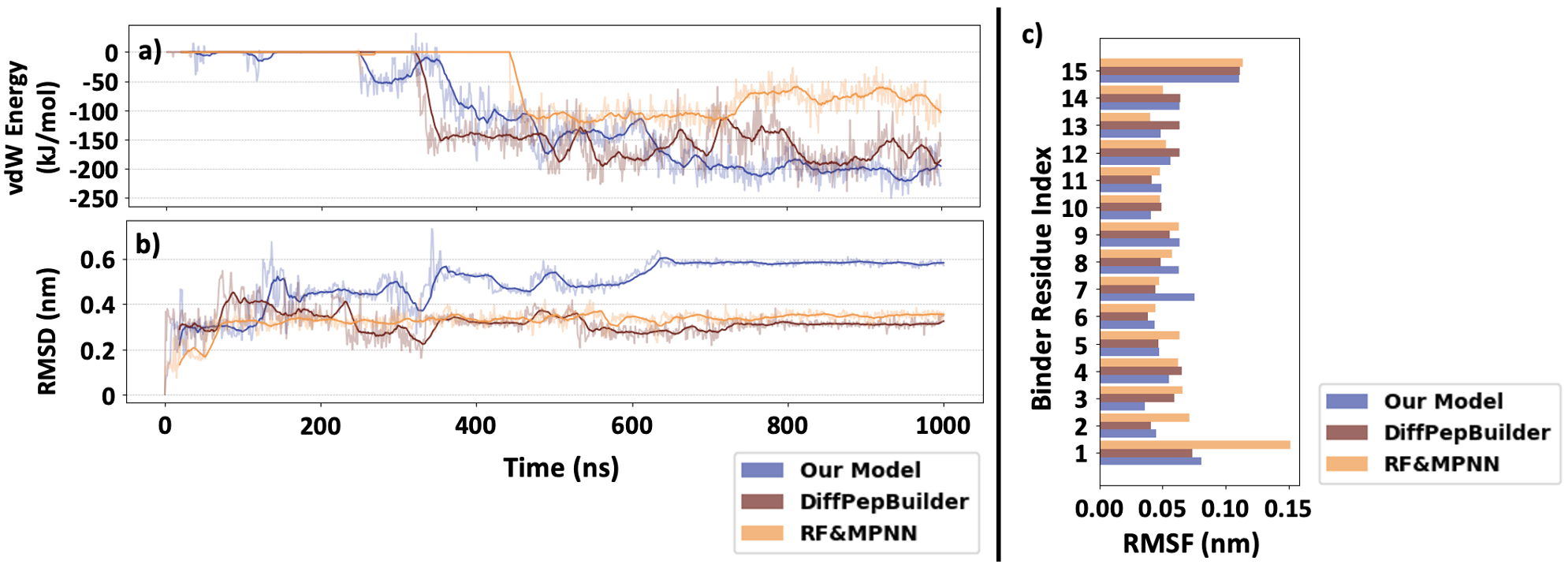}
    \caption{\textbf{Docking Simulation Analysis}. \textbf{a)} van der Waals (vdW) interaction energy between the peptide and TIGIT during the simulation. Our peptide exhibits the earliest interaction with TIGIT and also shows the strongest (lowest) interaction energy. \textbf{b)} Root-mean-square deviation (RMSD) of the peptide over the course of the simulation. All three peptides reach a stable conformation after approximately 800 ns. \textbf{c)} Root-mean-square fluctuation (RMSF) of the peptide throughout the simulation.}
    \label{fig:docking_simulation}
\end{figure}
\subsubsection{Dynamic Docking MD Simulation}
After selecting the top-performing binder from each method, we performed MD simulations to obtain a more accurate estimation of binding affinity. Figure \ref{fig:tigit_performance_str} shows the docking pose of three peptides. The peptide generated by our method shows the largest interaction interface towards TIGIT, as reflected by a $\Delta$ solvent-accessible surface area ($\Delta$SASA) of 1032.98~$\text{\AA}^2$, while the peptide from RF\&MPNN shows relative smaller region ($\Delta$SASA of $725.05$) and smallest from DiffPepBuilder ($\Delta$SASA of $718.90$). This indicate that our model has the ability to design peptides not only considering the pocket region but also the residues around it. For PepEDiff (our method), interactions between the peptide and pocket
residues were found in two out of three repeats, compared to only one repeat for each of the other two baseline
methods. Figure \ref{fig:docking_simulation}.a quantifies the interaction energy, showing that our peptide achieves a stronger (lower value) van der Waals (vdW) interaction of $-195.57$ kJ/mol towards TIGIT. DiffPepBuilder shows a slightly weaker interaction of $-185.40$ kJ/mol, followed by $-103.38$ kJ/mol from RF\&MPNN. Figure \ref{fig:docking_simulation}.b shows that all three peptides reach a stable conformation after approximately 800 ns. Figure \ref{fig:docking_simulation}.c shows that the peptide generated by the RF\&MPNN pipeline exhibits slightly higher fluctuations (average 0.07 nm) compared to 0.06 nm for both our model and DiffPepBuilder.

\begin{figure}[htbp]
    \vspace{-0.1 in}
    \centering
    \includegraphics[width=\columnwidth]{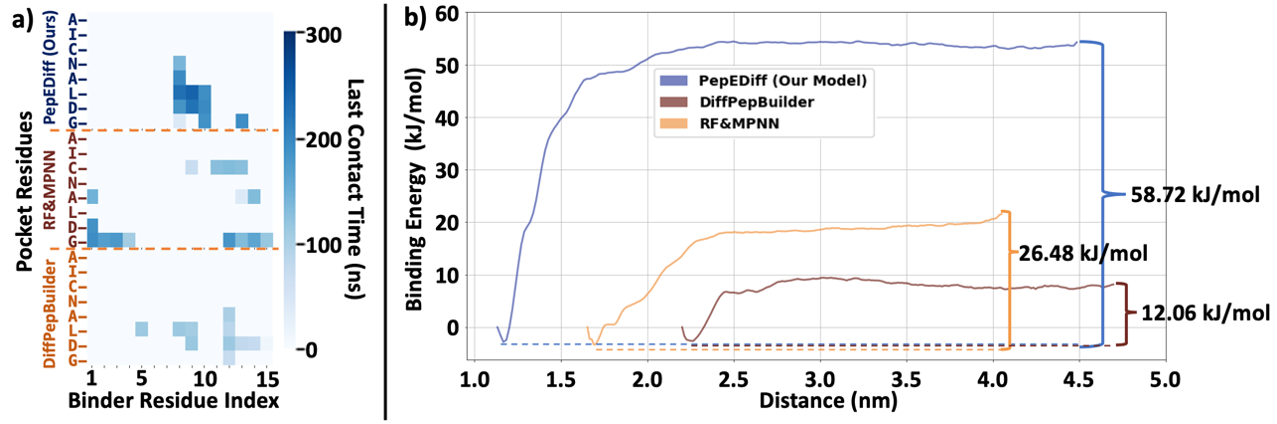}
    \caption{\textbf{Umbrella Sampling Analysis}.
    \textbf{a)}The heatmap (top) shows the last contact frame between binder residues and the pocket residues (A67–G74) during the pulling process. The line plot (bottom) displays the free binding energy estimated from umbrella sampling. \textbf{b)} A comparison of the binding energy for the three peptides.}
    \label{fig:umbrella_sampling}
\end{figure}

\subsubsection{Umbrella Sampling} The binding free energies estimated by umbrella sampling are shown in Figure~\ref{fig:umbrella_sampling}.b. The binder generated by our model exhibits the strongest binding energy of $58.72$~kJ/mol, while RF\&MPNN shows lower binding energy of $26.48$~kJ/mol followed by the lowest binding energy of $12.06$~kJ/mol from DiffPepBuilder. The Figure~\ref{fig:umbrella_sampling}.a illustrates the last contact time for each residue between pocket residues and the binders during the pulling phase of the umbrella sampling. Our model maintains its final contacts within the defined pocket residues. This suggests stronger and more stable engagement with the functional binding site. In contrast, the RF\&MPNN and DiffPepbuilder binders shows earlier dissociation from the pocket, indicating that its apparent binding affinity may primarily result from interactions with non-pocket residues rather than targeted binding at the functional interface.


\section{Conclusion}
In this work, we introduced a structure-independent peptide binder design framework that relies solely on receptor sequence and pocket residue information. Unlike existing methods that depend on structural modeling at the generation stage, our approach operates entirely in sequence space, enabling more effective optimization and better diversity of binder design. Through extensive evaluation on both general test sets and a case study targeting the immune receptor TIGIT, our model outperforms state-of-the-art baselines across key metrics, including binding affinity, sequence diversity, structure diversity, and even the embedding diversity.

Notably, through adopt latent exploration technique, our method exhibits higher diversity in both sequence and structure compared to prior approaches, highlighting its capacity to explore a broader binder space and potentially discover novel binding motifs. This diversity, combined with strong performance in binding energy evaluations, suggests that our embedding-based generation strategy effectively captures functional properties beyond explicit sequence similarity.

Despite these promising results, our approach has some limitations. First, although our method does not require structure input during generation, the evaluation of binding poses and energies still depends on downstream structure prediction models, such as Boltz-2\cite{passaro2025boltz} for structure prediction and Rosetta Commons toolkit\cite{chaudhury2010pyrosetta,leaver2013scientific} for fast relaxation. Second, while our current model focuses on general peptide binders, designing binders with more complex biochemical properties, such as cyclic structure or specific profiles, for example Lipinski's rule of five\cite{lipinski2012experimental}, remains an open challenge for future work. 

Overall, our results demonstrate the effectiveness and potential of sequence-based peptide binder generation, offering a promising direction toward more accessible and diverse therapeutic peptide design.


%
%
%
%
\begin{credits}

\subsubsection{\discintname}
The authors have no competing interests to declare that are
relevant to the content of this article.
\end{credits}
\newpage
\bibliographystyle{splncs04}
\bibliography{ref}
\end{document}


%

\title{Appendix}
%
%
\author{
Po-Yu, Liang\inst{1} \and Tibo, Duran\inst{2} \and Jun,  Bai\inst{1}\textsuperscript{\Letter}}
\institute{
    Department of Computer Science, University of Cincinnati, Ohio, United States\\
    \email{baiju@ucmail.uc.edu}
    \and
    Department of Microbiology and Plant Pathology, 
University of California, Riverside, California, United States
}
\maketitle              
\section{Model Implementation Details}

\begin{table}[h]\label{tab:hyperparameters}
\vspace{-1 em}
\centering
\caption{Training Hyper-Parameters}
\begin{tabular}{c|c}
    Batch Size & 8 \\
    Epoch & 500 \\
    Dropout Rate & 0.1 \\
    Learning Rate & $5*10^{-5}$ \\
    
    \hline
    \# of Hidden Layers & 2 \\
    Hidden Size & 2048 \\
    Intermediate Size & 4096 \\
    \# of Attention-Head & 8 \\
    \hline
    Diffusion Timestep & 1000 \\
\end{tabular}
\label{tab:angle_dihedral_def}
\end{table}

Our proposed model is an attention-based diffusion model that uses the BERT architecture as its backbone. The full set of model hyperparameters is provided in Table~\ref{tab:hyperparameters}, and the implementation code is available in our GitHub repository (https://github.com/LabJunBMI/PepEDiff-A-Peptide-binder-Embedding-Diffusion-Model).

\subsection*{Diffusion Framework}
We employ a time-dependent score-based diffusion model following \cite{song2020score}. 
Timestep information is encoded using Gaussian random Fourier features as described in \cite{tancik2020fourier}. 
For the diffusion process, we adopt the cosine variance schedule introduced by \cite{nichol2021improved}.

\subsection*{Optimization}
Training employs a linear warmup schedule over the first 10\% of epochs, following the recommendation in the original Transformer paper \cite{vaswani2017attention}.

\subsection*{Feature Normalization}
Input features (binder sequence embeddings) are normalized using a z-transform. The normalization statistics are computed exclusively on the training set and reused during inference for both the test set and the TIGIT binder generation task.

\subsection*{Embedding Model}
For ProtT5 embedding model, we use the pretrained \texttt{prot\_t5\_xl\_uniref50} variant. This version exhibits the strongest downstream performance based on the benchmarks reported on the ProtTrans\cite{elnaggar2021prottrans} repository (https://github.com/agemagician/ProtTrans) .

\section{Molecular Dynamics (MD) Simulation Details}

\subsection*{Overall Simulation Settings}
All molecular dynamics simulations were performed using \textsc{GROMACS}~2023\cite{abraham2015gromacs} with theLiquid Simulations All-Atom (OPLS-AA) force field\cite{jorgensen1996development}. The solvent environment consisted of extended simple point charge (SPC/E) water\cite{berendsen1987missing,chatterjee2008computational} with NaCl at a concentration of 0.137~mol/L to mimic physiological conditions. Temperature coupling was handled using the V-rescale\cite{bussi2007canonical} thermostat with a reference temperature of 300~K, while pressure coupling employed the Parrinello--Rahman\cite{parrinello1981polymorphic} barostat with a reference pressure of 1~bar.

\subsection*{Peptide Structure Preparation}
Peptide structures were generated using AlphaFold2\cite{jumper2021highly} rather than directly adopting predicted structures from upstream models. For each peptide, all five AlphaFold2 model variants were used to generate candidate structures, and the structure with the highest confidence score was selected. 

The selected peptide structure was placed in a cubic simulation box of size $4 \times 4 \times 4$~nm containing the solvated environment described above. The system was first minimized via energy minimization, followed by 1000~ps of NVT equilibration and 1000~ps of NPT equilibration. 

After equilibration, three independent 1000~ns production simulations were carried out. The final frame from each trajectory was extracted, and the structure with the smallest pairwise root-mean-square deviation (RMSD) to the other two was selected as the representative peptide structure.

\subsection*{Dynamic Docking Simulations}
To simulate peptide binding, a larger cubic simulation box of size $10 \times 10 \times 10$~nm was constructed containing the representative peptide structure and TIGIT (PDB ID: 3Q0H\cite{PDB_3Q0H}, chain A). The peptide was placed near the TIGIT binding pocket with an initial center-of-mass separation of approximately 30~\AA.

Energy minimization, followed by 1000~ps of NVT and 1000~ps of NPT equilibration, was performed as before. Three independent dynamic docking simulations were then run. The trajectory with the largest number of peptide--TIGIT contact residues (defined using a 4~\AA\ cutoff) within the pocket motif was selected for subsequent umbrella sampling.

\subsection*{Umbrella Sampling for Binding Free Energy Estimation}
Umbrella sampling simulations were performed to accurately estimate peptide--TIGIT binding free energies using the weighted histogram analysis method (WHAM)\cite{kumar1992weighted}. To reduce computational cost, the system was reduced to a $7 \times 10 \times 7$~nm box containing the peptide--TIGIT complex from the final frame of the selected docking trajectory.

During pulling, positional restraints (1000~kJ/mol/nm\textsuperscript{2}) were applied to the backbone atoms of ten TIGIT residues that do not contact the peptide. These residues differed among peptides based on docking geometry and were selected to allow the binding interface to remain fully flexible.

Pulling was performed along the peptide–TIGIT center-of-mass (COM) vector using a force constant of 650~kJ/mol/nm\textsuperscript{2} and a pulling rate of 0.009~nm/ps. Sampling windows were chosen based on the COM distance: 0.05~nm spacing for distances $< 2$~nm, 0.1~nm for distances between 2--4~nm, and 0.2~nm for distances $> 4$~nm. This adaptive window resolution ensures dense sampling in regions of strong interaction while improving computational efficiency at larger separations. Each umbrella sampling window was simulated for 10~ns prior to WHAM analysis.

\newpage
\bibliographystyle{splncs04}
\bibliography{ref}